\documentclass[linenumbers]{aastex631}

\usepackage{makecell}

\begin{document}
\nolinenumbers


\title{A New cross-domain strategy based XAI models for fake news detection

v0.1.1\footnote{Released on Feb 1st, 2023}}

\author{Deepak Kanneganti 20380423}
\affiliation{Curtin University \\
Bentley, Perth, WA, Australia,6152}

\begin{abstract}
\nolinenumbers

The Advancement in technology and rapid usage of social media has made communication easier and faster than ever before. Fake news threatens the community, democracy, egalitarianism and people's trust. Cross-domain text classification is a task of a model adopting a target domain by using the knowledge of the source domain. Natural Language Processing and Deep Learning models are used to identify misleading information. Explainability is crucial in understanding the behaviour of these complex models. In this study, we propose a four-level cross-domain strategy to study the impact of explainability on cross-domain models. The latest findings in the natural language process, the "Bidirectional Encoder Representations from Transformers" (BERT) model published by  ~\cite{devlin2018bert} google used to implement the concept of transfer learning. A fine- tune BERT model is used to perform cross-domain classification. Using this model, we conducted four experiments using datasets from different domains. Explanatory models like Anchor, ELI5, LIME and SHAP are used to design a novel explainable approach to cross-domain levels. The experimental analysis has given an ideal pair of XAI models on different levels of cross-domain.

\end{abstract}


\keywords{Bidirectional Encoder Representations from Transformers (BERT) --- Local Interpretable Model-Agnostic Explanations (LIME) --- Anchor --- Explain like I'm 5 (ELI5) --- Shapely Additive explanations (SHAP), Transformers , Tokenizers}


\section{Introduction} 

Nowadays, social media has become a potential influencing tool. According to the statistics published by Datareportal in July 2022, there is exponential growth in social media platforms, declaring that more than half of the world's population (59 per cent) is using them. Consequently, these platforms have deterministic effects on people's lives and the integrity of societies and local communities. Groups of people forming social media clusters use, unfortunately, these tools to spread speculation - so-called "fake news". In 2008, a journalist posted a report about Steve jobs medical condition. Many people retweeted this news. It has created massive confusion and controversy within societies and led to fluctuations in the stock price of Apple Inc. ~\cite{rubin2017deception}. During the Covid-19 pandemic, fake news was largely spread among people and has created panic within societies. Recent statistics published by the United States support receiving reports from 80 per cent of consumers about the fake news outbreak. Insufficient data is one of the reasons behind unreliable communication, making it difficult to distinguish fake from real news. In 2016, fake news was popular mainly during the United States elections. They have created a great source of influence on people's opinions about two constants. People made 171 million tweets in five months during the election. Statistics support ~\cite{allcott2017social}  that over 25 per cent of them are fake. During the Covid-19 pandemic, rumours on the internet have put the world under massive confusion about the disease's spread, vaccination and prevention. Numerous incidents are reported, showing that people even tried injecting bleach to protect themselves. 

It is an example of the potential effects of social media on people's opinions and acts. To deal with future challenges, efficiently encounter the impacts of fake news, and ensure reliable communication on social media platforms, the government and organizations need to implement systems that can detect the rumours and put the correct information in place. This paper uses data related to medical, political and financial sectors to develop a fine-tuned BERT model and a machine learning model for better performance in the cross-domain.

Explainability is an emerging domain in which organizations and business owners try to find explanations and solutions for problems with high detailed levels. Fake news often spread on social media starts to be unavoidable. The proposed solution from the cross-domain using BERT-Based text classification can help identify fake information regardless of the field. Explainability, crucial in social media platforms, government agencies, and the corporate world is considering implementing this solution before passing it to the decision making.This paper focuses on cross-domain text analysis by overlapping data from different domains. Explainable models are used to derive explanations of model performance.

\section{Contribution} \label{sec:stylew}

The consistent improvement in the usage of social media platforms shows a clear threat to governments and organizations. Thus, this paper aims to develop a fine-tuned BERT text classification model that works for the cross-domain and develop local agnostic Explainable models.

\begin{enumerate}
   \item List of contributions
   \begin{itemize}
     \item Designed a four-level (Low to High) Cross domain strategy to study the impact of explainable models
     \item Four explainable models are studied to drive the explanation of model performance.
     \item Identified two ideal models that give the best explanation using a five-level cross-domain strategy.
     \item Implemented two new explanatory models ELI5 and SHAP, in four different cross-domain models without changing the structure of the model.
     \item  A Potential improvement was highlighting the cross-domain using explainability
     \end{itemize}
   \end{enumerate}

\section{Related works} \label{sec:floats}
\subsection{Fake-news detection}

Fake news is used as a weapon by groups aiming to influence citizens' thinking and beliefs, thus creating discord in societies, which can hardly avoid in the internet era. The study of ~\cite{lazer2018science} supports that this is a concern of organizations all over the globe, suggesting the importance of the implementation of new integrated systems which can help detect misalliances in social media. According to the survey~\cite{zhou2018fake}, cascade-based and network-based techniques are the primary techniques used by previous authors to detect fake news. Cascade base detection distinguishes fake news using informative representation and computes the similarities between real and fake news. Network-based detection ~\cite{zhou2019network} recognized the fake by contributing the networks like homogeneous, heterogeneous and hierarchical to capture false information using the veracity scores. Linguist base method is a consistent approach to network base detention. This method focuses on identifying the information's grammar, syntax, readability and sentimental features to identify fake news. An instance of the Linguist based model is presented by ~\cite{choudhary2021linguistic}.
~\cite{10.5555/1953048.2078186} provide a path-breaking way to represent the works as vectors. The vectorizer contains various methods Bag of words, FT-IDF, GloVE and Word2Vec. Another study was presented by ~\cite{thota2018fake}, where the author discussed the implementation deep learning model using various vectorization methods to identify fake news.

The study on abnormal detection by~\cite{10.1145/3374749} gives an outcome applicable to detecting fake news. The study used the concepts of multi-layer perception (MLP) to identify abnormal behaviour, which can apply to this task.

\subsection{Cross domain text classification }

The concept of cross-domain is to develop a model that classifies the text of the target domain by using the knowledge of the source domain~\cite{li2012topic}. The aim is to reduce the gap between the source and target domains.
~\cite{pan2010cross}Pressed an approach to bridge the gap between the domain's spectral feature alignment algorithm(SFA) and line up the domain words from different fields into the cluster. These cluster helps to reduce the gap between the domain-specific words in two domains, which can use to reduce the gap between the two domains.
The multi-domain text classification approach presented by the author ~\cite{li2012multi}shows promising results in predicting the text belonging to a different domain. The author proposed a novel multi-domain active learning model that trained data instances from all disciplines, considering the duplicated information.

The following approaches may not have been designed for  fake news detection but could be successfully applied to perform cross-domain Structure corresponding learning (SCL) is a method by selecting a set of standard features from both sources and the targets domain using specific criteria like the highest degree of association, which has shown better results on cross-domain sentiment analysis, according to the author ~\cite{bollegala2015cross}

\subsection{Transfer learning}

Transfer learning is a method where pre-trained deep learning modes can perform similar NLP tasks on another dataset. The survey by ~\cite{pan2009survey} shows the progress of transfer learning in solving, machine learning models classification and regression. They discussed the future implementation of the domain adaptive, multitask learning and selection bias. The latest findings in the natural language processing BERT introduced by ~\cite{devlin2018bert}. The concept of transfer learning is critical in designing the BERT model., and the BERT model can perform pretraining and fine-tuning. Pretraining represents the process of training the general purpose of the language and the understanding of the model on a significant text source like wiki and books. Hugging Face has introduced various pre-trained models, which can help downstream NLP tasks. This paper presents an uncased BERT-based model for developing the fine-tuned model. The uncased BERT-based model is pre-trained on Book Corpus, a massive dataset containing eleven thousand unpublished books and English Wikipedia articles. The text knowledge gained through pretraining will help to understand the internal connections and relationships between the two kinds of writing. This pre-trained model can generate the random tokens according to the input parameters like max token, length and padding. In this study, the pre-trained BERT-based model will take id inputs and attention mask id as inputs to the fine-tuned model.

\subsection{Fine-tuned BERT model architecture and training} \label{sec:floats}

The pre-trained uncased BERT-based model is used to tokenize the text developed by Hugging Face~\cite{wolf2019huggingface}. The input sentence is tokens. Unique token [CLS] and [SEP] tokens are added to each sentence's beginning and ending. [CLS] is used to classify the task, and [SEP] represents the end of the sentence. Finally, the pre-training model provides a set of IDs to replace the tokens.
 
The following parameters are assigned to tokenize the text data. The max sentences with a length of 15, pad to max, and truncation as "True". A detail about the BERT architecture is explained by (). Uncased BERT-based model converts into the lower case before the tokenizing step.
 
The neural network with an input layer of 768 used a rectified linear unit (Relu) as the activation function. Dropout will reduce the effect of the model from over-fitting in a study where the dropout is 0.3. The final layer has two units as output with the soft-max function values. Adam Optimizer reduces the loss during back-propagation through the network, with a learning rate of 2e-05. The training loss and validation loss are evaluated through the parameter to analyze the model's performance. The neural network output gives the soft-max values of both classes, and the argmax function will classify the values as real or fake.

The concept of the fine-tuned BERT model has already been implemented in detecting fake news. The paper published by ~\cite{geetha2021improving} showed the performance of the pre-trained BERT model used in this regard. In this study, researchers performed experiments using several machine models and achieved good precision using BERT, with the model being fine-tuned with a soft-max layer for the classification task. Other researchers performed comparative analysis by combining other machine learning models and fine-tuning model development, which gives potentially better results.

\subsection{Explainable Models} \label{sec:style}
\subsection{Local Interpretable Model-agnostic Explanations (LIME)} \label{sec:style}

LIME proposed by ~\cite{ribeiro2016should} is the simplest way to give a meaningful interpretation. It is a model agonist method to give a local interpretation of the model behaviour in the neighbourhood of the sample. The main objective of the model is to derive meaningful explanations of complex models as they are linear at the local scale. Theoretically, they proved good local approximations. In this study, an interpretable linear mode is trained on 500 samples to predict and represent the sample's behaviour. The ideal concept suggests that every linear model can provide an excellent local estimation, and this is shown in the paper of ~\cite{ribeiro2016should} explaining the corresponding mathematical equation. The mathematical equation explained by the author is mentioned below.
\begin{center}
    $  f(x) = argminL(f,g,\pi x)+Omega(g) $
\end{center}

The objective is to find balance between minimization of $L(f , g , \pi x )$ and  human understandable levels of omega. Where g represented as potentially interpretable model.
$L(f,g,\pi x)$ quantifies how unfaithful g is in approximating explained model.

\subsection{Anchor}
The second explanatory model used in this analysis is Anchor~\cite{ribeiro2018anchors}, which is an algorithm based on the "if -then -else" rule. This model establishes the connection between features contributing to the model's decision. The anchor can show the relationship between words in a given sentence in this analysis.The mathematical formula of the anchor explained by the author ~\cite{ribeiro2018anchors}.The objective is explain a F(x) gives non-transparent model f : X → Y , X is the instance.X must be perturbed using a “perturbation distribution

\subsection{SHAP}
SHapley Additive exPlanations (SHAP) is a framework developed by the author ~\cite{lundberg2017unified} to determine why the model is unpleasantly drawing certain conclusions. Deep network connections are very complex to interpret. When decisions are crucial, the SHAP explanatory model helps analyse the results contributing to the model's behaviour. SHAP framework works on the concept of game theory. The SHAP explanatory model measures the relative contribution of each feature towards the final prediction. The critical metric to consider for the SHAP explanatory model performance is the base value and F(x). According to the paper of ~\cite{lundberg2017unified}, the base value E[f(x)] is E(y'). The mean value is displayed when there is no feature for the current output. F(x) is the expected value for the observation x. These are the critical parameters used to understand the SHAP prediction. SHAP plots can give a better understanding of the knowledge domain. The output of the plot's features is in red and blue colours. The red represents the features that push the prediction higher towards the right, and the blue represents the opposite effect. 

\subsection{ELI5}
According to ~\cite{fan2019eli5}, ELI5 is an exploitable model that can better explain several complex machine-learning algorithms. ELI5 stands for the term" I am 5". This explanatory model can show the weight of each prediction for the highlighted words and give score values for each given instance—a set of positive and negative features re-highlighted to support the decision of the prediction.

\section{Technology Stack}

Pandas and Numpy modules are the most commonly used for data science and machine learning tasks ~\cite{oliphant2006guide}. Pandas provide various options to perform data wrangling, whereas Numpy is used to create arrays and matrices. In each stage of the experiment with the explanatory models, these two modules play a crucial role. The predicted probability is a critical function that converts the model's output according to the explanatory model's requirements.

Transformers provide state of the art for architecture of the natural language, leading to a better understanding of the BERT model. The pre-trained uncased-BERT-based model and tokenizers are both from transformers of Hugging Face~\cite{wolf2019huggingface}. It is an open-source library that provides API and other tools to download the pre-trained models.

According to ~\cite{abadi2016tensorflow}, Tensor flow is the latest end-to-end open-source platform that provides libraries for high-level machine learning tasks. On the other hand, there is Kears, a neural network library that runs on Tensor flow, and these are two sources used to develop the neural network. The model requires a GPU to perform multiple computations, and CUDA helps to achieve computer graphical processing.

Sci-kit-learn is an open source that can perform useful statistics and machine learning operations like cross-validation, normalization and model development. ~\cite{pedregosa2011scikit}.

The book~\cite{gianfagna2021model} discusses the configuration of Explainable models.LIME library supports the black box classifier that supports Sci-kit-learn, and Keras ~\cite{ribeiro2016should}.Alibi library 
~\cite{klaise2021alibi}, is used in this study to design Anchor explainable model. A SpaCy, an open-source library used to train the pipelines as a part of developing the Anchor model. It can perform advanced Natural Language Processing (NLP) tasks ~\cite{vasiliev2020natural}. SHAP explainable algorithm is available in the SHAP package ~\cite{lundberg2017unified}. ELI5 ~\cite{gianfagna2021model}is a python package that supports Sci-kit-learn and Keras models to prevent bugs in machine learning models and to explain the model's prediction.

\section{Cross Domain Study Design}
The proposed cross-domain design strategy requires datasets from various domains. \textbf{Table 1} describes the datasets used to design the cross-domain research framework. This strategy used covid -19 data discussed in ~\cite{wang2020cord}. The dataset available in this study is a combination of text information from Twitter, kaggle and government organizations. In level 1, the model gets trained on the covid-19 domain, and it contains 17.8 thousand samples of Covid-19 data. The data combines an equal number of real and fake news. Mixed data is a second dataset category and a combination of the text sampled from various domains. For example, it combines domains like Political, Business, News articles, finance, weather and global news data. The Cross-domain dataset contains 9000 real and fake samples used in levels 2 and 3 experiments—it is similar to the mixed dataset with covid 19 data samples added to the data to evaluate the level of the cross-domain the model can perform in levels 2 and 3.

\begin{table}[!ht]
    \centering
    \begin{tabular}{|l|l|l|l|}
    \hline
        Data set & Sample size & Real & Fake \\ \hline
        Covid-19 & 17800 & 8900 & 8900 \\ \hline
        Mixed data & 9000 & 4500 & 4500 \\ \hline
        Cross domain & 40000 & 20000 & 20000 \\ \hline
    \end{tabular}
\caption{List of different domains datasets used in the experiments}
\label{tab:caption}
\end{table}

The primary objective of this study is to explore the explainable models in settings where the machine learning model predicts the samples in cross-domain configuration. Cross-domain refers to the work where the developed model classifies the text of the target domain by using the knowledge of the source domain. The domain can be completely different or vary in some parts. This section systematically discusses the various cross-domain configurations formulated to study the four explainable models. The configurations are designed from level 1 to level 4 based on the distribution gap between the source domain (training data) and the target domain (testing data). The gap is minimum for level 1 cross-domain study and maximum for level 4.

\begin{figure}[htp]
\centering
\includegraphics[width=0.8\textwidth]{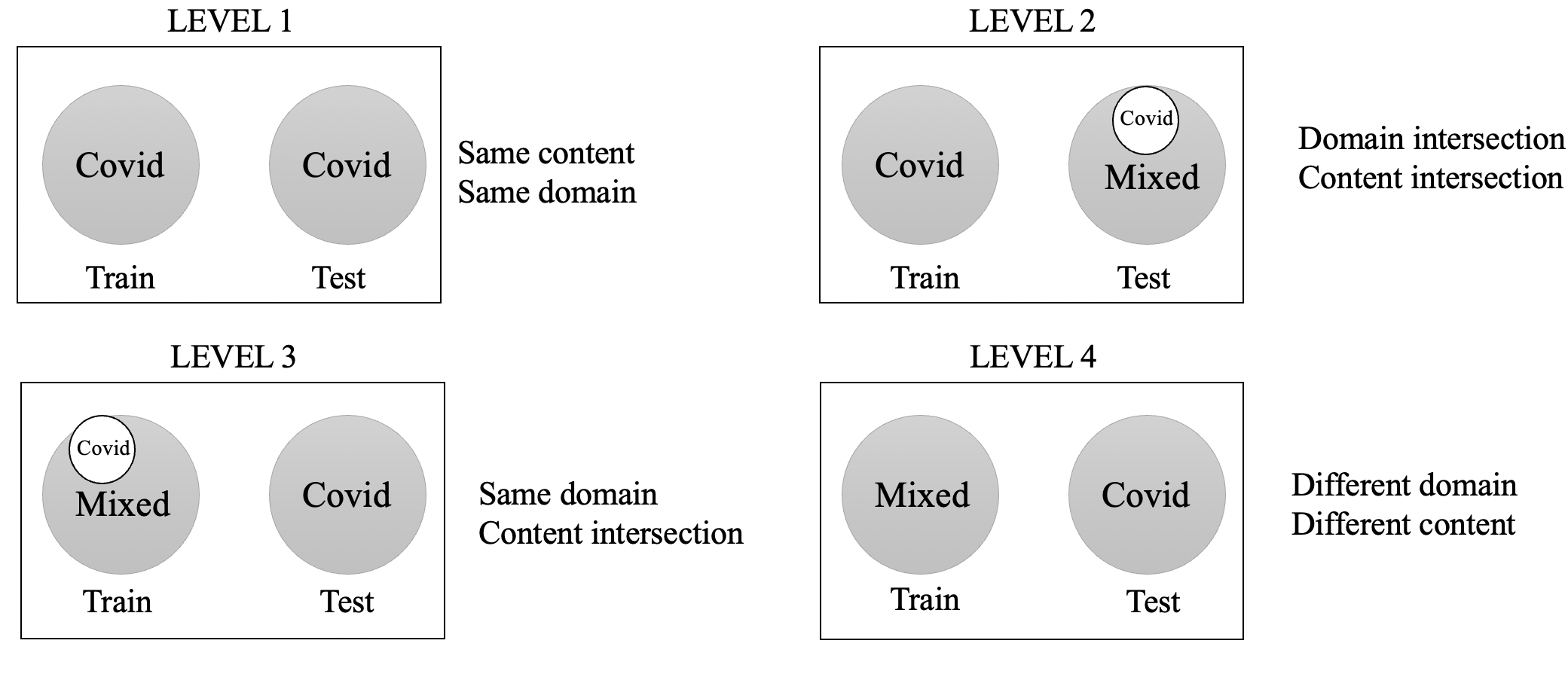}\hfill
\caption{Overlapping data for cross domain experiments}
\label{fig:figure3}
\end{figure}

\section{Experiments and Results} \label{sec:intro}

\textbf{Fig 1} gives a visual representation on the datasets designed to study the explainable models in the different cross-domain configurations. In the case of the lowest level 1, the domain of the training and testing set is the same. However, the testing samples are different from the training samples.COVID data discussed in ~\cite{wang2020cord} is used for experiment.In level 2, the distribution gap between the source and target domain is further increased by training the model on the COVID dataset and testing it on data from multiple domains (COVID, political, etc.). However, the source and target dataset still have some relations as the target dataset contains samples from COVID on which the model got trained.The source domain expands to the next level (i.e., level = 3), and the model gets trained on data from multiple domains (COVID, politics, etc.). However, the testing domain is constant, i.e. COVID. The objective is to study the generalizing capability of explainable models when the model gets trained on multiple domains. The maximum distribution gap for level 4 datasets with no direct relation between source and target domain. The source and target domain share only contextual relationships, as the samples in both domains are drawn from news articles.

Explainability is the second stage. The model's behaviour at various stages of the cross-domain strategy is derived using the XAI models like LIME, Anchor, ELI5, and SHAP.\textbf{Figure 2} gives a visual representation of the overview strategy. The data discus in \textbf{Table 1} is divided and trained on the model for all four levels of the cross-domain.

\begin{figure}[htp]
\centering
\includegraphics[width=0.8\textwidth]{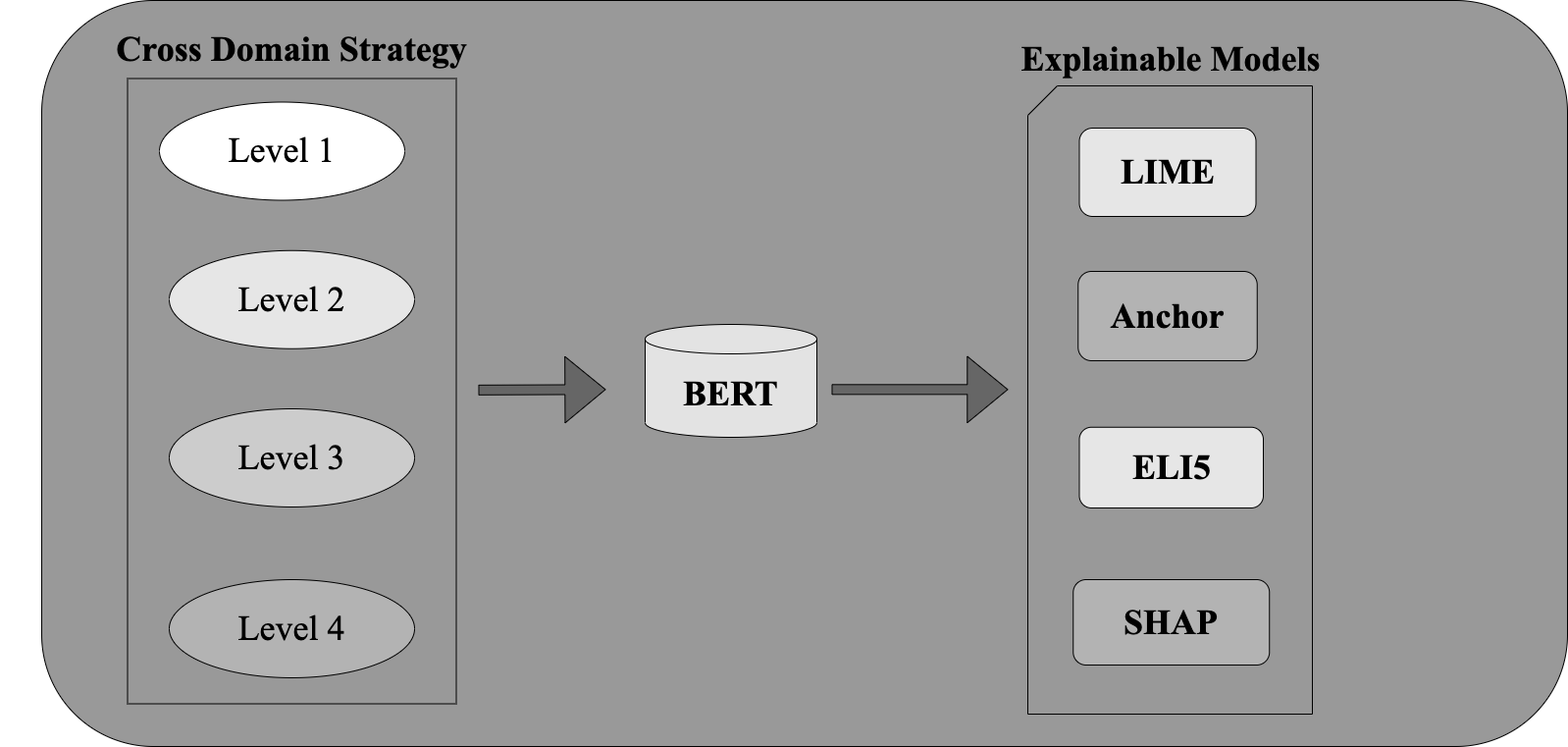}\hfill
\caption{The overview of the proposed approach}
\label{fig:figure3}
\end{figure}

\subsection{Cross Domain Strategy Evaluation}

\subsection{Accuracy}
Accuracy is the most crucial factor in measuring performance, and it refers to the ratio of correctly predicted observations to the total observations.

\begin{center}
\textbf{Accuracy: TP+TN/TP+FP+FN+TN}  
\end{center}

\subsection{Precision}
It is the ratio of correctly predicted positive observations in three classes. 

\begin{center}
\textbf{Precision = TP/TP+FP}
\end{center}

\subsection{Recall (Sensitivity)}
The recall is the ratio of correctly predicted positive observations of all classes. In a given dataset, there are three classes.
\begin{center}
\textbf{Recall = TP/TP+FN}
\end{center}
\subsection{F1-score}
F1 Score takes both false positives and false negatives into account. It is the weighted average of precision and recall and describes the model's quality.
 \begin{center}
\textbf{F1 Score = 2*(Recall * Precision) / (Recall + Precision)}
\end{center}

\subsection{ROC-AUC Curve}

AUC - ROC curve is a performance measurement used to evaluate actual and fake news classification.ROC is a probability curve, and AUC represents the degree or measure of separability. It gives a between picture to understand the model performed in differentiating classes.The higher value of AUC shows that better perform of model in classifying the real and fake news.

\section{List of experiments and results} \label{sec:intro}

The objective of these experiments is to analyze the performance of fine tuned BERT model in different domains.
\begin{enumerate}
   \item List of experiments
   \begin{itemize}
     \item BERT model training on the same domain, using Covid-19 data
     \item Model traning and learning evaluation using combined mixed dataset and Covid-19 content 
     \item Model training on Covid-19 data and teasing on mixed dataset 
     \item Model training on mixed data and evaluation on Covid-19 data for the cross domain 
\end{itemize}
\end{enumerate}

Achor SPACY USEAGE 
LIME 



%

\begin{table}[!ht]
    \centering
    \begin{tabular}{|l|l|l|l|l|l|}
    \hline
        Test cases & Accuracy & Label & Recall & Precision & F1 score \\ \hline
        Level 1 & 0.84 & Real & 0.84 & 0.84 & 0.84 \\ \hline
        ~ & ~ & Fake & 0.83 & 0.84 & 0.83 \\ \hline
        Level 2 & 0.68 & Real & 0.45 & 0.84 & 0.59 \\ \hline
        ~ & ~ & Fake & 0.91 & 0.62 & 0.74 \\ \hline
        Level 3 & 0.60 & Real & 0.95 & 0.57 & 0.71 \\ \hline
        ~ & ~ & Fake & 0.28 & 0.77 & 0.41 \\ \hline
        Level 4 & 0.57 & Real & 0.30 & 0.64 & 0.41 \\ \hline
        ~ & ~ & Fake & 0.83 & 0.54 & 0.65 \\ \hline
    \end{tabular}
    
\caption{Classification results of the four experiments using fine tuned BERT-Model}

\label{tab:caption}
\end{table}

\begin{table}[!ht]
    \centering
       \begin{tabular}{|l|l|l|l|l|}
    \hline
        Cases & Train data & Test data & Experiment & Best Explainability \\ \hline
        Level 1 & Covid-19 data & Covid-19 data & Tested Covid-19  Random Covid-19 & LIME,Anchor \\ \hline
        Level 2 & Mixed data & Covid-19 data & \multicolumn{1}{p{4cm}|}{Tested Covid-19, Random Covid-19 and Political data} & SHAP ,ELI5\\ \hline
        Level 4 & Covid-19 data & Mixed data & Random mixed data & LIME ,ELI5 \\ \hline
        Level 5 & Mixed & Covid-19 & Tested Covid-19  & LIME ,SHAP\\ \hline
    \end{tabular}
\caption{List of experiments and results}
\label{tab:caption}
\end{table}

\section{Experiments} \label{sec:intro}

\begin{figure}[htp]

\centering
\includegraphics[width=.3\textwidth]{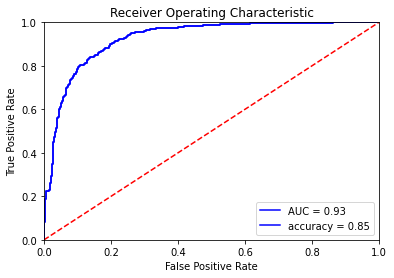}\hfill
\includegraphics[width=.3\textwidth]{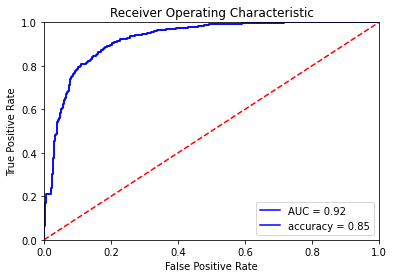}\hfill
\includegraphics[width=.3\textwidth]{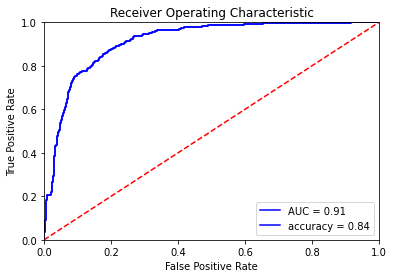}

\caption{AUC - ROC Curve for Experiment 1}
\label{fig:figure3}

\end{figure}

\label{sec:style}
\subsection{Explanation for level 1  of the common domain text classification model}

In case 1 Kaggle dataset on fake Covid-19 news is used to train the model. The dataset contains 17.8 thousand samples of Covid-19 data. The data is divided into three parts for training, validating and testing the model. The model is trained on 10.7 thousand samples and validated and tested on 3.4 thousand samples—a balanced dataset with an equal number of samples of real and fake news. The developed model for this case has performed constantly on all three stages with an accuracy of 85 per cent. AUC-ROC curve is a performance measurement used to evaluate problems of classification. It represents the ability of the curve to distinguish between the classes. \textbf{Figure 3} shows that the model is 93 per cent accurate for the differentiation between fake and real news.

\begin{enumerate}
   \item These are the list of explanations discussed below:
   \begin{itemize}
     \item True positive: when the BERT model is able to predict the real news correctly
     \item True negative: when the news are real but the model  predicted them as fake
     \item False positive: when the BERT model is able to predict the fake news correctly
     \item False negative: when the news are Fake but the model predicted them as real
     \item 3 Random Test cases to analyse the explainability
     \end{itemize}
   \end{enumerate}

In the experiment using the BERT model for detecting fake news about Covid-19, the model is 85 per cent accurate. LIME, Anchor, SHAP and ELI5 explanatory models help to explain the model's performance behaviour. LIME enables the interpretation of the model's behaviour along with the weight determination of each feature and its contribution to the model's decision-making. It is a local interpolation model designed to predict the feature using the sample instance around the given instance and can also be used to explain text data. In this experiment, the Local liner model was trained on 500 samples around the given feature. The explainable model shows the weights of the top 5 words and the probabilities of the decision. \textbf{Tables 4 and 5} show the results using LIME for experiment 1, each row for four possible model outcomes along with three random text data. The other rows in the table contain prediction probabilities, highlighted words and weights exhibiting the impact on the prediction probability. The "-" sign indicates that the chance of the sentence being real will increase, while the "+" sign indicates the increase in the chance of fake prediction.

\begin{table}[!ht]
    \centering
    \begin{tabular}{|l|l|l|l|}
    \hline
        Test cases & Text & P(Real) & P(Fake) \\ \hline
        True Positive & Covid Chinese health authorities confirm zero patient sex bats & 0.93 & 0.07 \\ \hline
        True negative & \multicolumn{1}{p{9cm}|}{Taking hot bath will prevent new coronavirus diseases taking hot bath}  & 0.10 & 0.90 \\ \hline
        False positive & Bananas can prevent new covid infections & 0.26 & 0.74 \\ \hline
        False negative & Cocaine used covid protection & 0.61 & 0.29 \\ \hline
        Random 1 & \multicolumn{1}{p{9cm}|}{Injecting or consuming bleach or disinfectant kills the virus} & 0.30 & 0.70 \\ \hline
        Random 2 & \multicolumn{1}{p{9cm}|}{Doctors in New-York city say the rate of kidney injury among covid is putting a strain} & 0.95 & 0.05 \\ \hline
        Random 3 & \multicolumn{1}{p{9cm}|}{Experts say the false negatives in covid tests probably occur due to an insufficient collection}  & 0.93 & 0.07 \\ \hline
    \end{tabular}
\caption{LIME explanations for the test case scenarios of experiment 1}
\label{tab:caption}
\end{table}

The table with Anchors outputs shows a row for each of the first four scenarios and the text of the used sentence. The example best explains precision. For instance, in the third row, precision is equal to 0.96. It is then probable that the perturbed example of the sentence "Trump looms behind both" is 0.96 high.

\begin{table}[!ht]
    \centering
    \begin{tabular}{|l|l|}
    \hline
        Highlighted words & Weights \\ \hline
        Authorities,  Zero ,Sex , Covid, Confirm, Chinese &  - 0.24,- 0.18, -0.16, +0.16, -0.15,-0.14 \\ \hline
        Prevent, Bath, Hot, Will, Disease &  +0.17, -0.14,+0.13, -0.07, +0.07 \\ \hline
        Bananas, Infections, Covid,  can , prevent & -0.25 ,+0.11,-0.05,  +0.041,-0.02 \\ \hline
        Used, Cocaine , Protection, Covid & -0.50,  -0.35,   +0.05,+0.21 \\ \hline
         Injecting, Disinfectant, or Consuming bleach & +0.17,+0.12,-0.10,+0.08, -0.06 \\ \hline
        Say, Rate, Doctor, Kidney, Infection  & -0.20,+0.14,-0.09,-0.08,+0.05 \\ \hline
        Experts, Test, Say, Occur, negative & -0.30,-0.23,-0.19,+0.13,+0.08 \\ \hline
    \end{tabular}
\caption{Highlighted words and weights of LIME explanatory model for experiment 1}
\label{tab:caption}   
\end{table} 

\textbf{Table 6} contains the Anchor output results. Precision is a metric used to explain the outcome. For instance, the precision is 0.96 for the true-positive scenario for the sentence "Covid Chinese health authorities confirm". It shows that declarations with the words "Chinese", "confirm," and "sex" will be classified as real news. The Anchor model uses the "if-then" rule to establish the connection between words. \textbf{Table 6} presents the highlights of the words and weights. The majority of the words have negative signs, which indicates that the prediction will be real. The words "Chinese", "confirm," and "sex" have negative weights, which supports the prediction further.

For case 2, the sentence "taking hot bath water will prevent new corona viruses", the model has correctly classified it as fake news. Since the model's performance is consistent in predicting both real and fake news, the Anchor shows 0.99 precision in this instance. In \textbf{Table 6}, LIME weights show a majority of positive signs, especially for the highlighted words "prevent", "Hot", and "disease".

The fourth case is an exciting outcome of understanding the model's behaviour. In the text" "Cocaine used covid protection", the Anchor was able to predict the text as fake with precision value of 1. The word "cocaine" and "protection" were expected as fake. For LIME, the output is the opposite, it predicts the text as actual news, and it is 60 per cent confident in the outcome. The fourth row in the \textbf{Table 6} has the highlighted words with equal numbers of positive and negative sign words. Anchor successfully predicts and finds the "if-else-then" rule for better explanations.

In the case of random text check, the LIME model successfully predicts the fake news for the given text "Injecting or consuming bleach or disinfectant kills the virus". It shows a confidence of 70 per cent. The highlighted words "injection" and "disinfection" have positive signs resulting in being the real news, and the Anchor is providing similar results. On the other hand, in this sentence containing "injection", "disinfection", and "bleach", the model is confident to classify as fake news.
\begin{table}[!ht]
    \centering
    \begin{tabular}{|l|l|l|l|l|}
    \hline
        Test cases & Text & Prediction & Precision & Anchors\\  \hline
        True positive & \multicolumn{1}{p{5cm}|}{Covid Chinese health authorities confirm zero patients sex bats} & Real & 0.96 & Confirm AND Chinese AND sex \\ \hline
        True negative & \multicolumn{1}{p{5cm}|}{Taking hot bath will prevent new coronavirus diseases taking hot bath} & Fake & 0.99 & prevent \\ \hline
        False positive & \multicolumn{1}{p{5cm}|}{Bananas can prevent new covid infections} & Fake & 1 & Infections AND prevent \\ \hline
        False negative & Cocaine used covid protection & Fake & 1 & Cocaine AND used \\ \hline
        Random 1 & \multicolumn{1}{p{5cm}|}{Injecting or consuming bleach or disinfectant kills the virus}
        & Fake & 1 & \multicolumn{1}{p{4cm}|}{Disinfectant AND injecting AND bleach} \\ \hline
        Random 2 & \multicolumn{1}{p{5cm}|}{Doctors in new york city say the rate of kidney injury among covid patients is putting a strain} & Real & 0.98 & Doctors \\ \hline
        Random 3 & \multicolumn{1}{p{5cm}|}{Experts say the false negatives in covid tests probably occur due to a non sufficient sample} & Real & 0.98 & Experts AND tests AND false \\ \hline
    \end{tabular}
\caption{Anchors explanations for the test case scenarios of experiment 1 }
\label{tab:caption}
\end{table}

Explanations acquired from the new explanatory models for the four case scenarios and random data are in  \textbf{tables 8}. The table with ELI5 presents predictions, probabilities and scores for all the cases discussed above. The Eli5 package can display the prediction along with the words' weights. The intercept represented as BIAS.\textbf{Table 8} has the information of the test cases, the prediction, probability and Score for all the test cases used in this experiment. The algorithm perforce can be explained using the probability and score values. As mentioned in the previous section proposed expandable model.ELI5 model displays the highlighted words in the shade with red and green colours based upon the value. The model's objective is to predict the fake news from the text. The comments highlighted in green represent the prediction's positive contribution. In contrast, the colour red is opposite.\textbf{Score} is an additional parameter that represents the prediction that the Score value \textbf{greater than} 0 is Fake news and Value \textbf{less than} 0 represents real news. It is an accuracy score weighted by the cosine distance between the generated sample and the original document.

For case one, the ELI5 explanatory model has correctly classified the text as real news with a probability of 0.89 in the sentence " Covid Chinese". The score value is less than zero, supporting that the text is real news.\textbf{Table 7} gives a visual explanation that the following words "Chinese", "authorities", and "confirm" show a positive contribution to the prediction. The third case is an example of the model correctly predicting fake news. The text "Bananas can prevent new covid infections" is fake news, and the model is 0.97 confident in its prediction. The score value is positive, which supports that the outcome "prevent" and "infections" are the two words that support the perdition. The last case is an exciting example of the model classifying fake news as real, showing a probability of 0.87 and a negative score. The model considers the words "cocaine" and "used" as positive, which explains its decision.

\begin{table}[!ht]
    \centering
    \begin{tabular}{|l|c|l|l|l|}
    \hline
        Cases & Text & Prediction & Probability  & Score \\ \hline
        1 & \multicolumn{1}{p{6cm}|}{Covid Chinese health authorities confirm zero patients sex bats} & Real & 0.804 & -1.410 \\ \hline
        2 & \multicolumn{1}{p{6cm}|}{Taking hot bath will prevent new coronavirus diseases taking hot bath} & Fake & 0.683 & 0.766 \\ \hline
        3 & Bananas can prevent new covid infections & Fake & 0.974 & 3.629 \\ \hline
        4 & \multicolumn{1}{p{6cm}|}{Cocaine used covid protection} & Real & 0.849 & -1.580 \\ \hline
        5 & \multicolumn{1}{p{6cm}|}{Injecting or consuming bleach or disinfectant kills the virus} & Fake & 0.795 & 1.356 \\ \hline
        6 & \multicolumn{1}{p{6cm}|}{Experts say the false negatives in covid tests probably occur due to insufficient collection of sample} & Real & 0.917 & -2.405 \\ \hline
        7 & \multicolumn{1}{p{6cm}|}{Doctors in new york city say the rate of kidney injury among covid-19 patients is putting a strain} & Real & 0.843 & -1.681 \\ \hline
    \end{tabular}
\caption{ELI5 explanations for the test case scenarios of experiment 1}
\label{tab:caption}
\end{table}


The last three cases are examples of the model reacting with random data. The sentence "Injecting or consuming bleach" is fake news, and the model predicted it correctly. The model shows a positive score, giving a probability of 0.79. "Consuming", "bleach", "kill", and "virus" are the contributing words for predicting the sentence as fake news.

\begin{table}[!ht]
    \centering
    \begin{tabular}{|l|c|l|l|l|}
    \hline
        Cases & Text & Prediction & F(Input)  & Base value \\ \hline
        1 & \multicolumn{1}{p{6cm}|}{Covid Chinese health authorities confirm zero patients sex bats} &  Real & -1.54035 & 0.1540 \\ \hline
        2 & \multicolumn{1}{p{6cm}|}{Taking hot bath will prevent new coronavirus diseases taking hot bath} &   Fake & 1.7930 & 0.154  \\ \hline
        3 & Bananas can prevent new covid infections &  Fake & 0.8451 & -0.23182 \\ \hline
        4 & \multicolumn{1}{p{6cm}|}{Cocaine used covid protection} &  Real & -1.152 & -0.4230 \\ \hline
        5 & \multicolumn{1}{p{6cm}|}{Injecting or consuming bleach or disinfectant kills the virus} &  Fake & 2.49306 & 0.1540 \\ \hline
        6 & \multicolumn{1}{p{6cm}|}{Experts say the false negatives in covid tests probably occur due to insufficient collection of sample} &Real & -3.008 & 0.15407 \\ \hline
    \end{tabular}
\caption{SHAP explanations for the test case scenarios of experiment 1}
\label{tab:caption}
\end{table}

The last case is an example of the model identifying real news on a random text. The sentence "experts say" refers to real information, and the words "experts" and "say" have positively contributed to the prediction. The model showed a high probability of 0.95 and a positive score, reflecting strength in the model performance.

Prediction using SHAP is analyzed using the parameters \textbf{base value} and \textbf{F(X)} value of each observation. The base value is E(y'), the mean value that the model will predict without feature. 
\textbf{Tabel 8} provides a visual explanation of the contribution of the words. The words highlighted in red and blue represent the features that enable the model to predict real news, whereas features with red refer to the features that permit the model to predict fake news. The real news "covid Chinese health authorities" is predicted as real by the model as most features are in blue. The F(x) value is negative since the x is real news. Both explainers have commonly highlighted words like "Chinese", "authorities", "confirm," and "bat". SHAP plot shows a majority of features in red colour with a positive predicted value (0.845) for the sentence "Bananas can prevent", concluding that the model has predicted the fake news correctly.



The fourth case shows abnormal results, where the model predicts fake news as accurate. The figure shows almost the exact contribution of the features of both types. The word "covid" was read separately as "cov" and "id", which confused the model. As a result, the model prediction for both random Covid-19 data. The model's performance was affected in predicting real and fake news. The last two columns in Tabel 7 show that the word "infection" has a majority of features contributing to a fake prediction. In contrast, the sentence "experts say" has a majority of features highlighted in blue, concluding that the model predicted the news as real.

\subsection{Explanation for level 2  on Mix domain training and Covid-19 domain testing }

In this scenario, the model trained with 32.4 thousand samples of a mixed dataset from various political, financial and Covid-19 domains. The model has been tested and validated on Covid-19 data to analyze the cross-domain performance. The model has been validated and tested with 17.4 and 6.8 thousand samples of Covid-19 data. The model shows good accuracy of 75 per cent for validation and testing. The model tested on covid 19 data; results showed that it better predicted fake news than real news. Table 2 has the results of case 2; the value of precision, recall and f1 score are less for real news, and the model is effective in predicting fake news. The \textbf{figure 4} shows the model evaluation metrics—the four explainable models are discussed further to interpret the model behaviour on different test case scenarios.

\begin{figure}[htp]
\centering
\includegraphics[width=.3\textwidth]{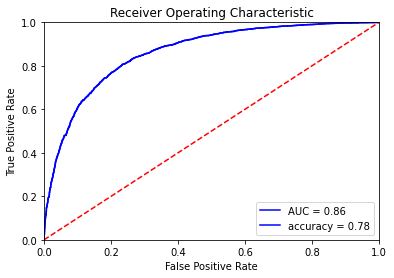}\hfill
\includegraphics[width=.3\textwidth]{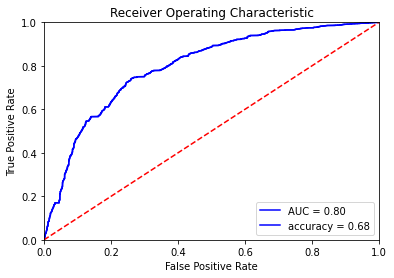}\hfill
\includegraphics[width=.3\textwidth]{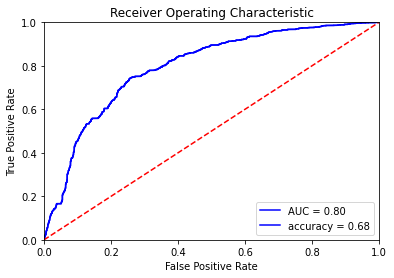}
\caption{AUC - ROC Curve for Experiment 2}
\label{fig:figure3}

\end{figure}

\begin{table}[!ht]
    \centering
    \begin{tabular}{|l|c|l|l|}
    \hline
        Test cases & Text & P(Real) & P( Fake) \\ \hline
        True positive & \multicolumn{1}{p{6cm}|}{Covid Chinese health authorities confirm zero patients sex bats} & 0.22 & 0.78 \\ \hline
        True negative & \multicolumn{1}{p{6cm}|}{Taking hot bath will prevent new coronavirus diseases taking hot bath} & 0.34 & 0.66 \\ \hline
        False positive & \multicolumn{1}{p{6cm}|}{Bananas can prevent new covid infections} & 0.26 & 0.74 \\ \hline
        False negative & Cocaine used covid protection & 0.30 & 0.70 \\ \hline
        Random Covid 1 & \multicolumn{1}{p{6cm}|}{Injecting or consuming bleach or disinfectant kills the virus} & 0.10 & 0.90 \\\hline
        Random Covid 2 & \multicolumn{1}{p{6cm}|}{Doctors in new york city say the rate of kidney injury among covid-19 patients is putting a strain} & 0.55 & 0.45 \\ \hline
        Random political 1 & \multicolumn{1}{p{6cm}|}{Osama Bin Laden older brother rents luxury student flats Scotland} & 0.47 & 0.53 \\ \hline
        Random Political 2 & \multicolumn{1}{p{6cm}|}{Trump looms behind Both Obama and Haley Speeches} & 0.58 & 0.42 \\ \hline
    \end{tabular}
\caption{LIME explanations for the test case scenarios of Experiment 2}
\label{tab:caption}
\end{table}

The explanation acquired from the LIME explainer for the four test cases and the random text data is in  \textbf{table 9 and 10}. The first and second contain the result of LIME, the probability of real and fake, highlighted words and weights. The table comprises anchor results with the same test cases, precision and anchored words. In test case one, the sentence "Covid Chinese health" is a sentence LIME model is predicting the text as fake news with a probability of 0.78. This exploration of the model predicting the real information as fake is due to the inefficiency of the model in identifying the real news; as discussed in the previous section, the result of case 2 shows that model is not efficient in predicting real news. The words "zero" and "Chinese" contribute to fake news.	The anchor is this test case showing the text with the words "patient", "conform", and "health" considered fake news and has the highest precision.

\begin{table}[!ht]
    \centering
    \begin{tabular}{|l|l|}
    \hline
        Highlighted words & Weights \\ \hline
        Sex ,Conform, Zero ,Health ,Chinese  & - 0.16,- 0.13,+ 0.13, - 0.05,+0.04 \\ \hline
        Prevent ,coronavirus ,Bath,taking,Hot & +0.24, -0.22,+0.17,-0.11,- 0.07 \\ \hline
        Covid,Bananes,prevent ,new,Infections &  -0.20 ,-0.30,+0.20,+0.02,-0.02 \\ \hline
        Used,Cocaine ,Protection ,covid & -0.06, -0.04, 0.02, 0.01 \\ \hline
        York,City,Kidney,Among ,injury & -0.12,-0.07,-0.05,-0.05,+0.04 \\ \hline
        Speeches,And,Looms,Trump,both & -0.13,-0.13,-0.07,-0.03,+0.01 \\ \hline
        Laden,Luxury,Flats,Reants,scotland & -0.09,-0.07,+0.05,-0.05,+0.03 \\ \hline
    \end{tabular}
\caption{Highlighted words and weights of LIME explanatory model for experiment 2}
\label{tab:caption}
\end{table}

As mentioned before, the anchors used in this study sometimes could not find the if-then—the third test case of the model clarifying correctly as fake news. The text for this case is "Bananas can prevent" anchor shows a prediction of 0.99 that the following news is fake. LIME explainer has the same outcome, with a 0.74 probability of fake text. 

The sentence "Cocaine used protection covid" was used in the last case. The previous model predicted this text as real news. LIME explainer was able to give a reasonable understanding with the words "Cocaine" and "used" words highlighted that are contributing to a fake prediction. The explainer shows a probability of 0.70 of the text being fake news. The anchor explainer could not capture the" if-then" rule.

\begin{table}[!ht]
    \centering
    \begin{tabular}{|l|l|l|l|l|}
    \hline
    Test cases & Text & Prediction & Precision & Anchors \\ \hline
        True positive & \multicolumn{1}{p{6cm}|}{Covid Chinese health authorities confirm zero patients sex bats} & Real & 1 & \multicolumn{1}{p{4.5cm}|}{patients AND confirm AND health} \\ \hline
        True negative & \multicolumn{1}{p{6cm}|}{Taking hot bath will prevent new coronavirus diseases taking hot bath} & Fake & 0.96 & prevent \\ \hline
        False positive & \multicolumn{1}{p{6cm}|}{Bananas can prevent new covid infections} & Fake & 0.99 & No anchor found \\ \hline
        False negative & Cocaine used covid protection & Fake & 1 & No anchor found \\ \hline
        Random Covid 1 & \multicolumn{1}{p{6cm}|}{Injecting or consuming bleach or disinfectant kills the virus} & Fake & 1 & virus \\ \hline
        Random Covid 2 & \multicolumn{1}{p{6cm}|}{Doctors in new york city say the rate of kidney injury among covid-19 patients is putting a strain} & Real & 0.99 & \multicolumn{1}{p{4.5cm}|}{Doctors AND York AND say AND kidney AND rate AND among} \\ \hline
        Random political 1 & \multicolumn{1}{p{6cm}|}{Osama bin laden older brother rents luxury student flats Scotland} & Fake & 0.96 & \multicolumn{1}{p{4.5cm}|}{older AND brother AND flats AND rents AND luxury AND Scotland} \\ \hline
        Random political 2 & \multicolumn{1}{p{6cm}|}{Trump looms behind Both Obama and Haley Speeches} & Real & 0.75 & speeches \\ \hline
    \end{tabular}
\caption{Anchors explanations for the test case scenarios of experiment 2}
\label{tab:caption}
\end{table}

The last three scenarios perform random text data from covid and the political domain. The first random text, "Doctors in New York", is real news. The model predicted it with a probability of 0.55, and the model could be more confident because the word in the sentence has almost the same contribution to real and fake. The words "York", "city", and "kidney" are positive words that support the real news. Anchor explainer performed well in identifying the if-then rule; it combined the words "doctors", "York", "say", "kidney"," rate", and "among". Mover-over, the precision is almost equal to 1; combining all these explanations, the model will predict the text which contains the following features.

The second random text relates to the political domain text "Trump looms behind". In this instance, the LIME explainer predicted correctly, as the text is real. With the words "speeches", "and", "looms" trump has the highest weight contributing to real prediction. According to the anchor explainer, the word "speech" is the anchor and shows a precision of 0.75 that the text is real.

The last text, "Osama bin laden older brother", is a fake news lime explainer, giving a probability of 0.53. In contrast, the anchor explainer has identified an if-then connection between the word "older", "brother", "flats", "rents"," luxury", and "Scotland". It has a precision of 0.96, and that text with these features gets classified as fake news.


Explanations acquired from the explanation models Eli5 are in \textbf{Table 12}. Each row represents the perdition, probability and score of the model. A score more excellent than zero represents fake news, and a negative score for real news. The first text, " Covid Chinese health", is real news; it is wrongly classified as fake by the model. The model shows a probability of 0.74 and a positive score. \textbf{Table 12} gives a better visual representation; the model identified that the negative words highlighted in red colour are inducing the model decision. The title " Cocaine uses" is fake news, and the model correctly classified it. 

In case 1, the model showed an inefficiency in identifying the fake news," Cocaine uses", but in this case, the model successfully identified it with a probability of 0.95 and a positive score of 3.053. The words "Cocaine" and "used" are highlighted in a dark colour

ELI5 explainer showed promising results in explaining the random covid data. The sentence "Injecting or consuming" and "Doctors in New York City say" are used to show the model predicting real and fake data correctly. The first text is fake news, the model categorized it with a probability of 0.70, and the score is 1.356.

\begin{table}[!ht]
    \centering
    \begin{tabular}{|l|l|l|l|l|}
    \hline
        Cases & Text & Prediction & Probability  & Score \\ \hline
        1 & \multicolumn{1}{p{6cm}|}{Covid Chinese health authorities confirm zero patients sex bats} & Fake & 0.764 & 1.173 \\ \hline
        2 & \multicolumn{1}{p{6cm}|}{Taking hot bath will prevent new coronavirus diseases taking hot bath} & Fake & 0.730 & 0.994 \\ \hline
        3 & \multicolumn{1}{p{6cm}|}{Bananas can prevent new covid infections} & Fake & 0.988 & 4.417 \\ \hline
        4 & \multicolumn{1}{p{6cm}|}{Cocaine used covid protection} & Fake & 0.955 & 3.053 \\ \hline
        5 & \multicolumn{1}{p{6cm}|}{Injecting or consuming bleach or disinfectant kills the virus} & Fake & 0.795 & 1.356 \\ \hline
        6 &\multicolumn{1}{p{6cm}|}{Doctors in new york city say the rate of kidney injury among covid-19 patients is putting a strain} & Real & 0.835 & -1.619 \\ \hline
        7 & \multicolumn{1}{p{6cm}|}{Osama bin laden older brother rents luxury student flats Scotland} & Fake & 0.661 & 0.670 \\ \hline
        8 & \multicolumn{1}{p{6cm}|}{Trump looms behind Both Obama and Haley Speech} & Real & 0.646 & -0.600 \\ \hline
        
    \end{tabular}
\caption{ELI5 explanations for the test case scenarios of experiment 2}
\label{tab:caption}
\end{table}

The random text belonging to the selected political domain is in the last two rows of the \textbf{Table 12}. According to the ELI5 explainer, the text " Doctor in New York" sample has an 85 per cent probability of being authentic news. The score value is negative -1.619, which gives more support to the outcome. The majority of the words are in green colour. The political text " Trump looms behind" shows a 64 per cent probability of being real news, and it gives a better understanding of the model outcome on real news. The text has very few shades of red colour highlighted words. The text "Osama bin laden brother" is predicted as fake with a probability of 65 per cent, as most words have a negative red shade.

\textbf{Table 13} shows the classification of the fake news for tested and random covid along with random political domain data. F(X) is a prediction value of the observation, and a positive value represents that the prediction is fake.

The instance "covid Chinese heath" shows equal red and blue highlighted words, f(x) value is 1.40 as a positive value, shows the explainer predicted it as fake news. The sentences "taking a hot bath" and "cocaine" are examples of correctly identified fake news, and both plots show a positive f(x) value. The words "prevent", "new", and "covid" are blue highlighted words. Along with another explainer, SHAP performed better on the instance "Cocaine", a misclassified example from the case1. 

Random covid data" expect says the false" is real news, and the model failed to predict accurately. The plot shows that the blue highlighted words infuse the prediction, but the value of f(x) is positive. Model identity the positive contribution words like "tests", "expects", "probably", and "negative", but till the f(x) and base value show the values of fake news. Overall SHAP model is inefficient in classifying the covid data on both tested and rand test cases.

\begin{table}[!ht]
    \centering
    \begin{tabular}{|l|l|l|l|l|}
    \hline
        Cases & Text & Prediction & F(Input)  & Base value \\ \hline
        1 & \multicolumn{1}{p{6cm}|}{Covid Chinese health authorities confirm zero patients sex bats} & Real & -1.404 & 1.9056 \\ \hline
        2 & \multicolumn{1}{p{6cm}|}{Taking hot bath will prevent new coronavirus diseases taking hot bath} & Fake &1.32204&1.9056 \\ \hline
        3 & \multicolumn{1}{p{6cm}|}{Bananas can prevent new covid infections} & Fake & 0.988 & 4.417 \\ \hline
        4 & \multicolumn{1}{p{6cm}|}{Cocaine used covid protection} & Fake & 2.570 & 1.254 \\ \hline
        5 & \multicolumn{1}{p{6cm}|}{Injecting or consuming bleach or disinfectant kills the virus} & Fake & 0.795 & 1.356 \\ \hline
        6 &\multicolumn{1}{p{6cm}|}{Experts say the false negatives in covid tests probably occur due to insufficient collection of sample} & Real & 0.9388 &1.9056 \\ \hline
        7 & \multicolumn{1}{p{6cm}|}{Osama bin laden older brother rents luxury student flats Scotland} &  Fake & 0.5096 & 1.722 \\ \hline
        8 & \multicolumn{1}{p{6cm}|}{Trump looms behind Both Obama and Haley Speech} &  Real & -0.7311 & 1.70997 \\ \hline
    \end{tabular}
\caption{SHAP explanations for the test case scenarios of experiment 2}
\label{tab:caption}
\end{table}

The random political text "Trump looms behind" and "Osama bin Laden" shows some promising results. The first test case is an example of correctly classified news real news; The blue colour features push higher toward the real prediction." Trump", "speeches", and "loom"  are the highlighted blue labelled words that contributed to the final prediction. The f(x) value is -0.73, which gives a better understanding of model predictions.

The other sentence, "Osama bin Laden" is fake news due to the positive value of F(x), and the words "older" and "student" are red labelled. The f(x) value is positive 0.93, contributing to a fake prediction. The plot shows that most of the features contribute to a real prediction.

\subsection{Explanation for level 3  on trained model on Covid-19 data and tested on mixed data}

In this scenario,  the model got trained on 23.4 thousand samples of covid dataset extracted from Twitter. The model is validated and tested with 2 thousand and thousand samples of data belonging to various political, financial and news domains. The model results show that the mode is 85 per cent on covid data and 61 per cent accurate on validation and test set. A close observation of the results in the \textbf{table 2} shows that the model is more accurate in predicting real news. The \textbf{figure 5} gives a visual understating of the model's performance in the third stage of the model development. These results show an average performance in predicting fake news from various domains using covid data. The model's accuracy in predicting mixed domain data is 60 per cent, and surprisingly the model shows less performance in predicting fake news than real news.

\begin{figure}[htp]
\centering
\includegraphics[width=.3\textwidth]{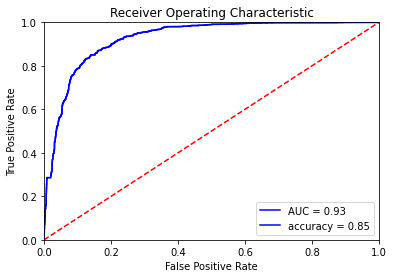}\hfill
\includegraphics[width=.3\textwidth]{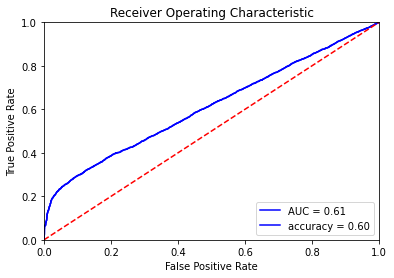}\hfill
\includegraphics[width=.3\textwidth]{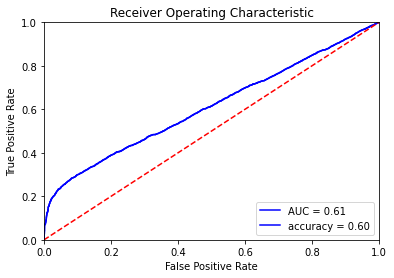}
\caption{AUC - ROC Curve for Experiment 3}
\label{fig:figure3}

\end{figure}

Results acquired from LIME and Anchor model from the 4 test scenarios and random text are in the \textbf{Table 14 and 15}. This table contains the following details: prediction probability, highlighted words and weights.\textbf{Table 16} shows the anchor explainer results for all the test cases.

\begin{table}[!ht]
    \centering
    \begin{tabular}{|l|l|l|l|}
    \hline
        Cases & Text & P(Real) & P( Fake) \\ \hline
        Mixed data political news 1 & \multicolumn{1}{p{6cm}|}{Trump campaign accidentally confirms that Trump illegally put money fidel castro hands} & 0.73 & 0.27 \\ \hline
        political 2 & Europeans future put in migrant crisis video & 0.49 & 0.51 \\ \hline
        political 3 & \multicolumn{1}{p{6cm}|}{Americans disconnected reality insouciant  become euphemism} & 0.53 & 0.47 \\ \hline
        political 4 & \multicolumn{1}{p{6cm}|}{Irrefutable engineered scientists lab are using well documented genetic engineering vector} & 0.30 & 0.70 \\ \hline
        Random Covid data  & 5G network spreads COVID -19 & 0.29 & 0.71 \\ \hline
        Random Covid data  & \multicolumn{1}{p{6cm}|}{Doctors in new york city say  the rate of kidney injury among covid-19 patients is putting a strain} & 0.71 & 0.29 \\ \hline
    \end{tabular}
\caption{LIME explanations for the test case scenarios of Experiment 3}
\label{tab:caption}
\end{table}

 The political data from the tested dataset and random covid are selected to draw a meaningful explanation of the model performance. The text "Trump campaign accidentally" is a correctly classified real news with a probability of 0.73, and the most impactful words for the model prediction are Trump, Fedel, illegally, conform and campaign. These words show negative weights, which is why the model predicts real news.
 
 The anchor explainer predicted the same result with a precision of 1. According to the anchor, the words campaign and confirm have an if-then rule, and the sentence with these words will be considered real news.
 
The sentence " Europeans future Putin migrant crisis video" is an example of an accurate fake prediction with a less confined probability of 0.51. The four majorly highlighted words share an equal positive and negative weights ratio. The word "Video" and "Putin" has negative weights, whereas words "European" and "future" shows positive. It shows that lime is unstable in dealing with the above example. 
The anchor model shows more confidence in predicting the exact text. The precision value is one, and the connection words are "future" AND "crisis" AND "Europeans" AND "migrant."

\begin{table}[!ht]
    \centering
    \begin{tabular}{|l|l|}
    \hline
        Highlighted words & Weights \\ \hline
        Trump, Fidel,illegally, Conform, campaign & -0.06,-0.05,-0.04,-0.02,-0.02 \\ \hline
        Viedo, Putin,Europeans,future &  -0.17,-0.15, 0.07, 0.05 \\ \hline
        Euphemism,insouciant ,americans, become ,disconnected & -0.20,0.08,-0.07,0.07,-0.02 \\ \hline
        Lab,Vector,Irrefutable,Well,genetic & 0.02,0.02,-0.02,0.01,-0.01 \\ \hline
        Spread, 19, Mobile ,5G & 0.03,0.01,0.00,0.001 \\ \hline
        Rate, In, Say, New-York, covid & 0.11,-0.10,-0.09,-0.07,0.06 \\ \hline
    \end{tabular}
\caption{Highlighted words and weights of LIME Explanatory model for Experiment 3}
\label{tab:caption}
\end{table}

The model needs to be more efficient in classifying real news. The sentence "Americans disconnected reality insouciant become euphemism" exemplifies the model classifying fake news as real. The LIME explainer shows a P(real) as 0.53. The words in the sentence have almost the same positive and negative contributions, resulting in a  less confident prediction. Anchor has mispredicted it with a precision value of 0.96. The model suggested that the words "euphemism" AND "Americans" influence the prediction of real news.

\begin{table}[!ht]
    \centering
    \begin{tabular}{|l|l|l|l|l|}
    \hline
        Test cases & \multicolumn{1}{c|}{Text} & Prediction & Precision & \multicolumn{1}{c|}{Anchors} \\ \hline
        True positive & \multicolumn{1}{p{6cm}|}{Trump campaign accidentally confirms trump illegally put money fidel castro hands} & Real & 1 & -  \\ \hline
        True negative & \multicolumn{1}{p{6cm}|}{Europeans future putin migrant crisis video} & Fake & 0.96 & \multicolumn{1}{p{3.8cm}|}{future AND crisis AND europeans AND migrant} \\ \hline
        False positive & \multicolumn{1}{p{6cm}|}{Irrefutable engineered scientists lab using well documented genetic engineering vector} & Fake & 0.95 & lab  \\ \hline
        False negative & \multicolumn{1}{p{6cm}|}{Americans disconnected reality insouciant become euphemism} & Real & 0.96 & euphemism AND americans \\ \hline
        Random 1 & 5G network spreads COVID -19 & Fake & 0.99 & spread \\ \hline
        Random 2 & \multicolumn{1}{p{6cm}|}{doctors in new york city say the rate of kidney injury among covid-19 patients is putting a strain} & Real & 1 & doctors AND say \\ \hline
    \end{tabular}
\caption{Anchors explanations for the test case scenarios of experiment 3}
\label{tab:caption}
\end{table}

According to the LIME explainer, the sentence "irrefutable engineered scientists" is fake news. The model miss-classified it as real with a probability of 0.70. The highlighted words show positive weights, which resulted in a wrong prediction. The anchor model showed only one word, "Lab", and the precision is 0.95. 

Random covid text" 5G network " and " Doctor in New York" are examples of real and fake news predictions. Both explainers showed good performance in predicting accurately. The most impacting words for the first text are " Spread", "19", "Mobile", and "5G", and they have positive weights. The second text has most words with negative weights, which shows a real prediction. The second explainer showed the anchor words "Spread " with a good precision of 0.99. The second text also has an if-then combination of the words "Doctor" AND "Say".

\textbf{Table 17} show the explanations obtained using the ELI5 model. The sentence  "Trump campaign accidentally" is correctly classified with a high confidence P(Real) 1. The model shows a negative score of -7.73. The majority of the words are in green colour.

The sentence "Europeans future Putin" is an example of the ELI5 explainer wrongly classifying fake news as Real. The model shows a high probability of 0.86, and the score value is negative. The visual plots show a mixed red and green colour, but the words with green colour dominated the other words in red.

\begin{table}[!ht]
    \centering
    \begin{tabular}{|l|c|l|l|l|}
    \hline
        Cases & Text & Prediction & Probability  & Score \\ \hline
        1 & \multicolumn{1}{p{6cm}|}{Trump campaign accidentally confirms trump illegally put money fidel castor hands} & Real & 1.00 & -7.624 \\ \hline
        2 & \multicolumn{1}{p{6cm}|}{Europeans future putin migrant crisis video} & Real & 0.868 & -1.557 \\ \hline
        3 & \multicolumn{1}{p{6cm}|}{Americans disconnected reality insouciant become euphemism} & Real & 0.752 & -1.109 \\ \hline
        4 & \multicolumn{1}{p{6cm}|}{Irrefutable engineered scientists lab using well documented genetic engineering vector} & Fake & 0.785 & -1.297 \\ \hline
        5 & \multicolumn{1}{p{6cm}|}{5G network spreads COVID -19} & Fake & 0.967 & 3.365\\ \hline
        6 & \multicolumn{1}{p{6cm}|}{Doctors in new york city say the rate of kidney injury among covid-19 patients is putting a strain} & Real & 0.785 & -1.297 \\ \hline
    \end{tabular}
\caption{ELI5 explanations for the test case scenarios of experiment 3}
\label{tab:caption}
\end{table}

The "Americans disconnected reality insouciant" is mislabeled as fake like all the other explainers. The performer of the model is less one of the reasons behind the wrong prediction. Most of the words highlighted in the fig for this instance are in green.

The last sentence is an explanation of random covid data. The sentence "5G network spreads COVID -19" is fake news, and explained obtained by the ELI5 is accurate. The results show a probability of 0.96 and a positive score of 3.36. "Doctor in New York City" was predicted as real news with a probability of 0.78. the words "doctor", "new", "York", and "say" were lighted as positive contributions.

The SHAP explainer used for this experiment is in \textbf{Table 18}. The instances "Trump comping" and "Europe future" are examples of real and fake news, and the SHAP explainer was successfully labelled correctly. The plot shows that most of the words in the sentence "trump campaign" are heightened in blue, contributing to a real label prediction. The words"Trump", "hands", "conform", "accidental", and "campaign" have high weight. In the fake prediction example, the words future, in, migrant and crisis are the red-labelled words that contribute to the fake prediction. f(x) is 0.277, a positive value that gave the fake precision.

\begin{table}[!ht]
    \centering
    \begin{tabular}{|l|c|l|l|l|}
    \hline
        Cases & Text & Prediction & F(Input)  & Base value \\ \hline
        1 & \multicolumn{1}{p{6cm}|}{Trump campaign accidentally confirms trump illegally put money fidel castor hands} & Real & -7.313 & 0.1372 \\ \hline
        2 & \multicolumn{1}{p{6cm}|}{Europeans future putin migrant crisis video} &real & 0.273 & -0.291 \\ \hline
        3 & \multicolumn{1}{p{6cm}|}{Americans disconnected reality insouciant become euphemism} & Real & -0.1948 & 0.0289 \\ \hline
        4 & \multicolumn{1}{p{6cm}|}{Irrefutable engineered scientists lab using well documented genetic engineering vector} & fake & 0.1372 &1.30085 \\ \hline
        5 & \multicolumn{1}{p{6cm}|}{5G network spreads COVID -19}  & Fake &2.32557 & -0.026065\\ \hline
        6 & \multicolumn{1}{p{6cm}|}{Doctors in new york city say the rate of kidney injury among covid-19 patients is putting a strain}& Real & -2.027 &0.1272 \\ \hline
    \end{tabular}
\caption{SHAP explanations for the test case scenarios of experiment 3}
\label{tab:caption}
\end{table}

The texts "Americans disconnected" and "irrefutable engineered scientist" are examples of the model prediction wrongly. The first sentence is fake news, and the figure shows that it has an almost equal number of weights highlighted in red and blue that infuse the prediction. The F(X) value is -0.198, which is harmful, showing that the model failed to predict it as fake. Another instance is an example of wrongly predicted real news as fake. Due to most of the words being in red colour, the explainer predicted it was fake. The words "Engineered lab well are the infusing words in the sentence.

The random covid sentence "5G mobile network spread COVID-19" is an instance of a model identifying fake covid data. The SHAP explainer identified the words "5G", "spread", and "Covid". The positive F(x) value adds more evidence to the explanation.

In the instance " Doctor in new york city says the", the SHAP explainer was reasonably sure to consider it real. The plot shows the words "Doctor", "say," are "york" were highlighted with blue, pushing towards the left. The value of the prediction is negative gives more support to the explanation.

\subsection{Explanation for level 4  trained on different domain and tested on different domain}

Experiment four is to drive the performance of the highest cross-domain. The model gets trained on 31 thousand samples of mixed data from various domains, and none of the data samples has covid 19 related data. To analyze the model's efficacy in detecting fake news on the cross-domain covid data set with 17.5 thousand sampled and 6.6 thousand samples used for validation and testing.\textbf{Fig 6}  is the ROC and AUC graphs for training, validation and test scenarios. The training model is 81 per cent accurate, and the testing set is 57 per cent accurate. The AUC - ROC curve is a performance measurement , the results shows the model is no very effect in differentiating classes.

\begin{figure}[htp]
\centering
\includegraphics[width=.3\textwidth]{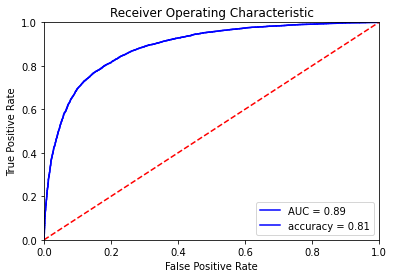}\hfill
\includegraphics[width=.3\textwidth]{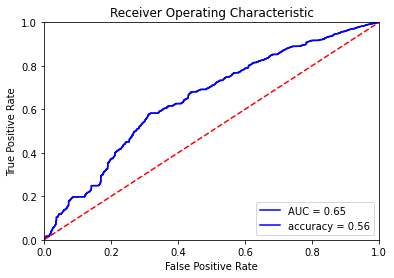}\hfill
\includegraphics[width=.3\textwidth]{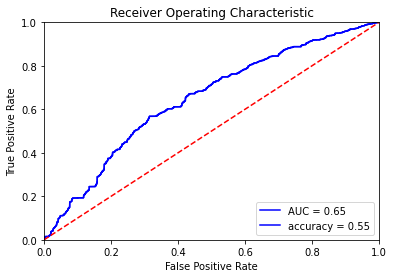}
\caption{AUC - ROC Curve for Experiment 4}
\label{fig:figure3}

\end{figure}

The first test scenario is the text "Covid Chinese health authorities", which is incorrect perdition. The real news is labelled as fake by the LIME and Anchor explainers. The first columns in \textbf{Table 19} show that the majority of the words in this text show positive weights. The anchor explanation failed to identify the if-then connection in the sentence. The sentence " Taking hot bath will" is an example of a model that exactly predicted the fake news since it shows good performance in identifying the fake news than the real. LIME explainer showed a probability of 0.67. The words "prevent" has the highest positive weight among all the other highlighted words. Anchor the similar perforce and label the sentence as fake with a precision of 0.99.

\begin{table}[!ht]
    \centering
    \begin{tabular}{|l|l|l|l|}
    \hline
        Test cases & \multicolumn{1}{c|}{Text} & P(Real) & P(Fake) \\ \hline
        True Positive & \multicolumn{1}{p{6cm}|}{Covid Chinese health authorities confirm zero patient sex bats} & 0.27 & 0.73 \\ \hline
        True negative & \multicolumn{1}{p{6cm}|}{Taking hot bath will prevent new coronavirus diseases taking hot bath}  & 0.33 & 0.67 \\ \hline
        False positive & \multicolumn{1}{p{6cm}|}{Bananas can prevent new covid infections} & 0.27 & 0.73 \\ \hline
        False negative & Cocaine used covid protection & 0.28 & 0.72 \\ \hline
        Random 1 & \multicolumn{1}{p{6cm}|}{Injecting or consuming bleach or disinfectant kills the virus} & 0.27 & 0.73 \\ \hline
        Random 2 & \multicolumn{1}{p{6cm}|}{Experts say the false negatives in covid tests probably occur due to an insufficient collection} & 0.28 & 0.72 \\ \hline
        Random 3 & \multicolumn{1}{p{6cm}|}{Trump looms behind both Obama and Haley speeches}  & 0.64 & 0.36 \\ \hline
        Random 4 &\multicolumn{1}{p{6cm}|}{ osama bin laden older brother rents luxury student flats scotland} & 0.49 & 0.51 \\ \hline
        
    \end{tabular}
\caption{LIME explanations for the test case scenarios of Experiment 4}
\label{tab:caption}
\end{table}

The sentence "Cocaine used covid protection" is incorrectly predicted as real news in the previous overlap cross-domain experiments. Surprisingly the highest cross-domain model could perform well in classifying it as real news. Anchor did succeed in finding an if-then rule, and the anchor combination is "Cocaine" AND "used". Moreover, the precision value is equal to 1; when the words appear in the sentence, the model will always predict it is fake news. The anchor for this section shows a precision of 1, and this sentence with the words "Infections" AND "prevent." will be considered fake news.

\begin{table}[!ht]
    \centering
    \begin{tabular}{|l|l|}
    \hline
        Highlighted words & Weights \\ \hline
       Zero, conform, Chinese, patients,authorities  &  -0.017,+0.01,+0.013,+0.009,+0.008\\ \hline
        Prevent, taking, coronavirus,new, bath & +0.15,-0.05,-0.05,-0.05,-0.04 \\ \hline
        Can, prevent,covid, bananas ,infection& +0.01,+0.0005,-0.0049,-0.003,0.003 \\ \hline
        used,covid,protection,cocaine & -0.004,+0.002,-0.001,+0.0005\\ \hline
         Injecting, disinfection,consuming,bleach,kills & +0.013,+0.006,-0.006,-0.005,+0.04 \\ \hline
        negatives,probably,insufficient,experts,say &-0.012,+0.09,+0.0009,+0.008,-0.008 \\ \hline
        and,speeches,Trump,looms,Obama & -0.13,-0.107,-0.09375,-0.08,+0.03 \\ \hline
        bin,flats,luxury,laden.student & -0.08,0.07,-0.06,-0.064,-0.06\\ \hline
        
    \end{tabular}
\caption{Highlighted words and weights of LIME explanatory model for experiment 4}
\label{tab:caption}   
\end{table} 

For Random test scenarios, the content "Injecting or consuming bleach" was the first case. In this instance, the LIME explainer shows a reasonable probability of 0.75 in considering it as fake. The words "injecting", "disinfection", and "kills." are contributed words for the prediction.

"Trump looms behind both Obama and Haley speeches" is random policy data used to explain the model's efficiency in predicting real data. LIME explained showed promising results in labelling are correct. The words "and", "speeches", "Trump", and "looms" shows good contributions in considering it as real news. The anchor model successfully identifies the if-then rule between the word "Speech" AND "looms".

\begin{table}[!ht]
    \centering
    \begin{tabular}{|l|l|l|l|l|}
    \hline
        Test cases & \multicolumn{1}{c|}{Text} & Prediction & Precision & \multicolumn{1}{c|}{Anchors}\\  \hline
        True positive & \multicolumn{1}{p{6cm}|}{Covid Chinese health authorities confirm zero patients sex bats} & Fake & 0.96 & - \\ \hline
        True negative & \multicolumn{1}{p{6cm}|}{Taking hot bath will prevent new coronavirus diseases taking hot bath} & Fake & 0.99 & \multicolumn{1}{p{2cm}|}{taking and prevent} \\ \hline
        False positive & \multicolumn{1}{p{6cm}|}{Bananas can prevent new covid infections} & Fake & 1 & \multicolumn{1}{p{2cm}|}{Infections AND prevent} \\ \hline
        False negative & Cocaine used covid protection & Fake & 1 & \multicolumn{1}{p{2cm}|}{Cocaine AND used} \\ \hline
        Random 1 & \multicolumn{1}{p{6cm}|}{Injecting or consuming bleach or disinfectant kills the virus}
        & Fake & 1 & Disinfectant \\ \hline
        Random 2 & \multicolumn{1}{p{6cm}|}{Experts say the false negatives in covid tests probably occur due to a non sufficient sample} & Real & 0.98 &  - \\ \hline
        Random 3 & \multicolumn{1}{p{6cm}|}{Trump looms behind both Obama and Haley speeches}  & Real & 0.96 & \multicolumn{1}{p{2cm}|}{Speech AND looms}\\ \hline
        Random 4 & \multicolumn{1}{p{6cm}|}{osama bin laden older brother rents luxury student flats scotland}  & Fake & 0.96 & \multicolumn{1}{p{2cm}|}{older and brother} \\ \hline
    \end{tabular}
\caption{Anchors explanations for the test case scenarios for experiment 4 }
\label{tab:caption}
\end{table}

Lastly, the model correctly classified random political news for the sentence' The Osama bin laden older brother rents luxury student flats Scotland' as fake news. The LIME  explanation shows that the model has assigned relatively similar probabilities between real and fake. The probability showed that the model is not confident in the prediction. Anchor showed reasonable confidence in labelling it as fake news with a precision value of 0.96. The highlighted words are "older" and "brother".

\begin{table}[!ht]
    \centering
    \begin{tabular}{|l|l|l|l|l|}
    \hline
        Cases & \multicolumn{1}{c|}{Text} & Prediction & Probability  & Score \\ \hline
        1 & \multicolumn{1}{p{6cm}|}{Covid Chinese health authorities confirm zero patients sex bats} & Fake & 0.96 & 3.171 \\ \hline
        2 & \multicolumn{1}{p{6cm}|}{Taking hot bath will prevent new coronavirus diseases taking hot bath} & Fake & 0.843 & 1.680 \\ \hline
        3 & \multicolumn{1}{p{6cm}|}{Bananas can prevent new covid infections}& Fake & 1.0 & 8.663 \\ \hline
        4 & Cocaine used covid protection & Fake & 1.00 & 8.121 \\ \hline
        5 & \multicolumn{1}{p{6cm}|}{Injecting or consuming bleach or disinfectant kills the virus}& Fake & 0.994 & 5.190\\ \hline
        6 & \multicolumn{1}{p{6cm}|}{Experts say the false negatives in covid tests probably occur due to a non sufficient sample} & Fake & 0.973 & 3.586 \\ \hline
        7 & \multicolumn{1}{p{6cm}|}{Trump looms behind both Obama and Haley speeches} & Real & 0.897 & -2.169 \\ \hline
        8 & \multicolumn{1}{p{6cm}|}{osama bin laden older brother rents luxury student flats scotland} & Real & 0.550 & -0.200 \\ \hline
    \end{tabular}
\caption{ELI5 explanations for the test case scenarios of experiment 4}
\label{tab:caption}
\end{table}

\textbf{Table 22} contains the results extracted using the ELI5 model for the four covid and random text test scenarios. Probability and score are the main parameters to understand the model explanation. The sentence "Covid Chinese health" shows a high confidence of 0.96 and a positive score of 3.171. The model failed to classify correctly and predicted the real news as fake. The text "Taking hot bath" and "Cocaine used" exemplify the model classifying fake news. For the first sentence, the plot \textbf{fig 12} shows that the words "coronavirus" and "talking" are the most infusing words for the fake prediction. The "Cocaine" sentence is an example of misclassified fake news from experiment one. The model showed evidence in labelling it as fake in the highest cross-domain.

Random covid data is in the fifth and sixth rows of table 22. The ELI5 showed good performance in reading the results. The text "injecting or consuming" represents correctly labelled fake news. The explainer showed a high confidence of 0.99 and a considerable score value of 5.190. The plot shows the words "Consuming" and "Bleach" are highlighted in dark red, which gives an understanding of the influence of these words on the prediction.

"Experts say the false negative" sentence is real news; the model failed to predict it correctly. The model shows a probability of 0.973 and a positive score. Overall, experiment four is the highest cross-domain; the model is 57 accurate in classifying fake news. The model shows more effect in classifying fake news than real for covid domain data.

Random political data samples are used to understand the model's efficiency in other domains. The text "Trump looms behind" is real news; the explainer predicted it with a probability of o.89 and a negative score of -2.169. Another text, " Osama bin Laden" is a fake sample, and the explainer predicted it as real with a minimal probability of 0.5450. The plot shows that most of the words are in green, positively impacting the prediction words.

\begin{table}[!ht]
    \centering
    \begin{tabular}{|l|l|l|l|l|}
    \hline
        Cases & \multicolumn{1}{c|}{Text} & Prediction & F(Input)  & Base value \\ \hline
        1 & \multicolumn{1}{p{6cm}|}{Covid Chinese health authorities confirm zero patients sex bats} & Fake & 4.86 & 1.92 \\ \hline
        2 & \multicolumn{1}{p{6cm}|}{Taking hot bath will prevent new coronavirus diseases taking hot bath} & Fake & 1.805& 1.927 \\ \hline
        3 & \multicolumn{1}{p{6cm}|}{Bananas can prevent new covid infections}& Fake & 5.022 & 1.55382 \\ \hline
        4 & Cocaine used covid protection & Fake & 4.01365 & 1.385  \\ \hline
        5 & \multicolumn{1}{p{6cm}|}{Injecting or consuming bleach or disinfectant kills the virus}& Fake & 4.2038 & 1.92812\\ \hline
        6 & \multicolumn{1}{p{6cm}|}{Experts say the false negatives in covid tests probably occur due to a non sufficient sample}& Fake &3.00935 & 1.92812  \\ \hline
        7 & \multicolumn{1}{p{6cm}|}{Trump looms behind both Obama and Haley speeches} & Real  & -1.3649 & 1.57607  \\ \hline
        8 & \multicolumn{1}{p{6cm}|}{osama bin laden older brother rents luxury student flats scotland}  & Real & 0.00709 & 1.595 \\ \hline
    \end{tabular}
\caption{SHAP explanations for the test case scenarios of experiment 4}
\label{tab:caption}
\end{table}

Lastly, the SHAP explainer results are in the \textbf{Table 23}. The first test case, "Covid Chinese health", is incorrectly classified as fake. Looking at the F(x) value, the model is unreliable, showing a high positive value of 4.86. The weights of the following words "Covid", "Conform", "Zero", and "bat" are pushing towards a fake prediction.

The news "Taking a hot bath will prevent" and "cocaine used "are examples of accurately classified as fake news. The plot shows that the model identified many words contributing to a fake prediction in both cases. Looking at the positive score provided by the SHAP, the model is confined to the prediction.

The third test case model was incorrect, and the real news was classified as fake. Looking at the probabilities provided by LIME, the model's uncertainty was minimal, and the assigned probability that the prediction is fake is equal to 95. While the weights of the particular words are relatively low, comparison with Anchors' output suggests the high impact of the names on the model's classification results. 

Random covid samples "Injecting or consuming" and, "Experts say the false negative" are real and fake examples. The explainer results show that the model needs to be more specific in classifying the data related to the covid domain. The efficiency in predicting fake news shows more promising results than real, and the model requires more performance knowledge and effectively identifying real news.

"Trump looms behind" and "Osama bin laden older" texts are for random political test scenarios. The first sentence is as real news. The plot shows that the words "speech", "trump", and "Hillary" in the sentence are pushing toward an accurate prediction. The value of f(X) is negative, which gives more evidence to the prediction.

Lastly, the model classified the sentence "Osama bin Laden" as fake news. The analysis of the SHAP explainers showed that words like "older" and "student" are critical, which outweigh the other words that can lead to a real news prediction. The low value of F(x) shows minimal model uncertainty.

\section{Conclusion}

This study explores four-level (low to high) cross-domain stars using the local interpreting model to identify the ideal pair that gives better explanations. Experiment one is a low-level cross-domain classification; the models LIME, Anchor, ELI5, and SHAP have shown exemplary performance in explaining. In experiment two, the model was trained on mixed data and tested on covid data; the results of LIME and ELI5 gave a good explanation. In the third experiment, a model was trained on covid domain data and tested on mixed data. ELI5 and Anchor are ideal parts in giving better explanations. The last experiment, the highest level of cross-domain LIME and SHAP models, showed consistency in classifying the news accurately. Overall the study gave good results in defining a pair of models that help to give a better explanation. This study develops new explainable models ELI5 and SHAP to fine-tune BERT without changing the model.

\section{Future work}

The study gave a better understanding of model efficiency on various stages of cross-domain. This objective is to develop a novel, explainable strategy using this series of experiments as a extension to this study.


\begin{acknowledgments}
The authors would like to thank Dr Sajib Mistry ,Dr Aneesh krishna and Dr. Nancy agarwal for their helpful insights, and for providing us with relevant code and/or data related to their projects.I would like to thank all people who have contributed to the elaboration of this paper.
\end{acknowledgments}

\vspace{5mm}
\facilities{HST(STIS), Swift(XRT and UVOT), AAVSO, CTIO:1.3m,
CTIO:1.5m,CXO}




\appendix


\bibliography{bibr.bib}{}

\begin{thebibliography}{}
\expandafter\ifx\csname natexlab\endcsname\relax\def\natexlab#1{#1}\fi
\providecommand{\url}[1]{\href{#1}{#1}}
\providecommand{\dodoi}[1]{doi:~\href{http://doi.org/#1}{\nolinkurl{#1}}}
\providecommand{\doeprint}[1]{\href{http://ascl.net/#1}{\nolinkurl{http://ascl.net/#1}}}
\providecommand{\doarXiv}[1]{\href{https://arxiv.org/abs/#1}{\nolinkurl{https://arxiv.org/abs/#1}}}

\bibitem[{Abadi {et~al.}(2016)Abadi, Barham, Chen, Chen, Davis, Dean, Devin,
  Ghemawat, Irving, Isard, {et~al.}}]{abadi2016tensorflow}
Abadi, M., Barham, P., Chen, J., {et~al.} 2016, Savannah, GA

\bibitem[{Allcott \& Gentzkow(2017)}]{allcott2017social}
Allcott, H., \& Gentzkow, M. 2017, Journal of economic perspectives, 31, 211

\bibitem[{Bollegala {et~al.}(2015)Bollegala, Mu, \&
  Goulermas}]{bollegala2015cross}
Bollegala, D., Mu, T., \& Goulermas, J.~Y. 2015, IEEE Transactions on Knowledge
  and Data Engineering, 28, 398

\bibitem[{Choudhary \& Arora(2021)}]{choudhary2021linguistic}
Choudhary, A., \& Arora, A. 2021, Expert Systems with Applications, 169, 114171

\bibitem[{Collobert {et~al.}(2011)Collobert, Weston, Bottou, Karlen,
  Kavukcuoglu, \& Kuksa}]{10.5555/1953048.2078186}
Collobert, R., Weston, J., Bottou, L., {et~al.} 2011, J. Mach. Learn. Res., 12,
  2493–2537

\bibitem[{Devlin {et~al.}(2018)Devlin, Chang, Lee, \&
  Toutanova}]{devlin2018bert}
Devlin, J., Chang, M.-W., Lee, K., \& Toutanova, K. 2018, arXiv preprint
  arXiv:1810.04805

\bibitem[{Fan {et~al.}(2019)Fan, Jernite, Perez, Grangier, Weston, \&
  Auli}]{fan2019eli5}
Fan, A., Jernite, Y., Perez, E., {et~al.} 2019, arXiv preprint arXiv:1907.09190

\bibitem[{Geetha \& Renuka(2021)}]{geetha2021improving}
Geetha, M., \& Renuka, D.~K. 2021, International Journal of Intelligent
  Networks, 2, 64

\bibitem[{Gianfagna \& Di~Cecco(2021)}]{gianfagna2021model}
Gianfagna, L., \& Di~Cecco, A. 2021, in Explainable AI with Python (Springer),
  81--113

\bibitem[{Klaise {et~al.}(2021)Klaise, Van~Looveren, Vacanti, \&
  Coca}]{klaise2021alibi}
Klaise, J., Van~Looveren, A., Vacanti, G., \& Coca, A. 2021, J. Mach. Learn.
  Res., 22, 181

\bibitem[{Lazer {et~al.}(2018)Lazer, Baum, Benkler, Berinsky, Greenhill,
  Menczer, Metzger, Nyhan, Pennycook, Rothschild, {et~al.}}]{lazer2018science}
Lazer, D.~M., Baum, M.~A., Benkler, Y., {et~al.} 2018, Science, 359, 1094

\bibitem[{Li {et~al.}(2012{\natexlab{a}})Li, Jin, \& Long}]{li2012topic}
Li, L., Jin, X., \& Long, M. 2012{\natexlab{a}}, 26, 998

\bibitem[{Li {et~al.}(2012{\natexlab{b}})Li, Jin, Pan, \& Sun}]{li2012multi}
Li, L., Jin, X., Pan, S.~J., \& Sun, J.-T. 2012{\natexlab{b}}, in Proceedings
  of the 18th ACM SIGKDD international conference on Knowledge discovery and
  data mining, 1086--1094

\bibitem[{Lundberg \& Lee(2017)}]{lundberg2017unified}
Lundberg, S.~M., \& Lee, S.-I. 2017, Advances in neural information processing
  systems, 30

\bibitem[{Oliphant(2006)}]{oliphant2006guide}
Oliphant, T.~E. 2006, A guide to NumPy, Vol.~1 (Trelgol Publishing USA)

\bibitem[{Pan {et~al.}(2010)Pan, Ni, Sun, Yang, \& Chen}]{pan2010cross}
Pan, S.~J., Ni, X., Sun, J.-T., Yang, Q., \& Chen, Z. 2010, in Proceedings of
  the 19th international conference on World wide web, 751--760

\bibitem[{Pan \& Yang(2009)}]{pan2009survey}
Pan, S.~J., \& Yang, Q. 2009, IEEE Transactions on knowledge and data
  engineering, 22, 1345

\bibitem[{Pedregosa {et~al.}(2011)Pedregosa, Varoquaux, Gramfort, Michel,
  Thirion, Grisel, Blondel, Prettenhofer, Weiss, Dubourg,
  {et~al.}}]{pedregosa2011scikit}
Pedregosa, F., Varoquaux, G., Gramfort, A., {et~al.} 2011, the Journal of
  machine Learning research, 12, 2825

\bibitem[{Ribeiro {et~al.}(2016)Ribeiro, Singh, \&
  Guestrin}]{ribeiro2016should}
Ribeiro, M.~T., Singh, S., \& Guestrin, C. 2016, in Proceedings of the 22nd ACM
  SIGKDD international conference on knowledge discovery and data mining,
  1135--1144

\bibitem[{Ribeiro {et~al.}(2018)Ribeiro, Singh, \&
  Guestrin}]{ribeiro2018anchors}
Ribeiro, M.~T., Singh, S., \& Guestrin, C. 2018, 32

\bibitem[{Rubin(2017)}]{rubin2017deception}
Rubin, V.~L. 2017, in The SAGE handbook of social media research methods
  (Sage), 342

\bibitem[{Thota {et~al.}(2018)Thota, Tilak, Ahluwalia, \&
  Lohia}]{thota2018fake}
Thota, A., Tilak, P., Ahluwalia, S., \& Lohia, N. 2018, SMU Data Science
  Review, 1, 10

\bibitem[{Tian {et~al.}(2020)Tian, Luo, Lu, Su, Sun, \&
  Zhang}]{10.1145/3374749}
Tian, Z., Luo, C., Lu, H., {et~al.} 2020, ACM/IMS Trans. Data Sci., 1,
  \dodoi{10.1145/3374749}

\bibitem[{Vasiliev(2020)}]{vasiliev2020natural}
Vasiliev, Y. 2020, Natural Language Processing with Python and SpaCy: A
  Practical Introduction (No Starch Press)

\bibitem[{Wolf {et~al.}(2019)Wolf, Debut, Sanh, Chaumond, Delangue, Moi,
  Cistac, Rault, Louf, Funtowicz, {et~al.}}]{wolf2019huggingface}
Wolf, T., Debut, L., Sanh, V., {et~al.} 2019, arXiv preprint arXiv:1910.03771

\bibitem[{Zhou \& Zafarani(2018)}]{zhou2018fake}
Zhou, X., \& Zafarani, R. 2018, arXiv preprint arXiv:1812.00315, 2

\bibitem[{Zhou \& Zafarani(2019)}]{zhou2019network}
---. 2019, ACM SIGKDD explorations newsletter, 21, 48

\end{thebibliography}
\bibliographystyle{aasjournal}



\end{document}